\documentclass[runningheads]{llncs}
\usepackage[utf8]{inputenc}
\usepackage{amsmath}
\usepackage{graphicx}
\usepackage{hyperref}
\usepackage{cleveref}
\usepackage{multirow}
\usepackage{float}
\usepackage{booktabs}
\usepackage{tabularx}
\usepackage{subfigure}
\usepackage{ulem}

\begin{document}
\title{Learning from students' perception on professors through opinion mining}
\titlerunning{Learning student's perception}
%

\author{Vladimir~Vargas-Calderón\inst{1}\orcidID{0000-0001-5476-3300} \and Juan~S.~Flórez \inst{1}\orcidID{0000-0003-3730-705X} \and Leonel~F.~Ardila\inst{1}\orcidID{0000-0002-8012-0066} \and Nicolas~Parra-A.\inst{1}\orcidID{0000-0002-1829-4399} \and
Jorge~E.~Camargo\inst{3}\orcidID{0000-0002-3562-4441} \and
Nelson~Vargas\inst{3}}
\authorrunning{V. Vargas-Calderón et al.}
%
\institute{Laboratorios~de~Investigación~en~Inteligencia~Artificial y Computación~de~Alto~Desempeño, Analista, Bogotá, Colombia\\ \email{\{vvargasc,jsflorezj,lfardilap,nparraa\}@analist.ai}\\
\url{http://www.analist.ai}\and
System~Engineering~Department, Fundación~Universitaria Konrad~Lorenz, Carrera~9~Bis~No.~62~-~43, Bogotá, Colombia\\
\email{\{jorgee.camargom,nelsona.vargass\}@konradlorenz.edu.co}}
\maketitle              
\begin{abstract}

Students' perception of classes measured through their opinions on teaching surveys allows to identify deficiencies and problems, both in the environment and in the learning methodologies. The purpose of this paper is to study, through sentiment analysis using natural language processing (NLP) and machine learning (ML) techniques, those opinions in order to identify topics that are relevant for students, as well as predicting the associated sentiment via polarity analysis. As a result, it is implemented, trained and tested two algorithms to predict the associated sentiment as well as the relevant topics of such opinions. The combination of both approaches then becomes useful to identify specific properties of the students' opinions associated with each sentiment label (positive, negative or neutral opinions) and topic. Furthermore, we explore the possibility that students' perception surveys are carried out without closed questions, relying on the information that students can provide through open questions where they express their opinions about their classes.




\end{abstract}
%
%
%
%
%
%
\section{Introduction}

Having a clear picture of students' perception on their classes, professors, and university facilities enables educational institutions to propose strategies to improve in many areas. It has been suggested by many studies that positive students' perception on the learning environment is correlated with higher academic achievement~\cite{samdal1999,ELHILALI2015420,Jamilian2013,diseth2007,Baek2002}. Therefore, not only can universities improve the quality of their professors, their class content as well as learning facilities, but they also can improve --as a consequence-- their students' academic achievement, leading to an overall improvement of the education quality.

The call for action is clear. However, in order to propose and implement effective improvement strategies, one needs to measure the students' perception. Typical ways of doing this is through evaluations carried out at the final stage of each academical period where students grade their professors in several aspects. These evaluations normally consist of an online questionnaire with closed questions, and some open questions where students give their opinions about the class and their professors. Closed questions questionnaire can be tedious for students, leading to low response rates~\cite{wright2017,JEPSON2005103,Adams1982,ROLSTAD20111101,Sahlqvist2011,smith2003effect,iglesias2000}. Closed questions are helpful for the fast interpretation of results with statistical tools. These questions are designed to measure professors' performance on specific topics such as how engaging the class is, punctuality, among others. On the other hand, open questions provide students with a free space to express their opinions. Of course, gathering and interpreting data from open questions responses is a much more challenging task than making statistics from closed questions. Nonetheless, the amount of useful information found in students' opinions is a valuable source that is rarely exploited.

The latest advances in machine learning and natural language processing (NLP) techniques can be used to build tools that facilitate the analysis of large amounts of opinions generated by students. Particularly, sentiment analysis is suited to identify and quantify how positively or negatively students feel about their professors. These machine learning applications have only been recently explored~\cite{dobre2015students}. For instance, Naïve Bayes has been used to classify students' opinions in social media~\cite{Permana_2017}. Also, Latent Dirichlet Allocation (LDA)~\cite{blei2003latent} has been used to model topics along with sentiment analysis to explore opinions from students~\cite{quteprints115064}. Some studies using tools from machine learning have been conducted in the field of students' perception analysis. The majority of them have addressed the issue of performing sentiment analysis of the students' comments~\cite{HEW2020103724,Skrbinjek2019}, and others have tried to identify topics in suggestions and opinions left by students ~\cite{Gottipati2018}, thus, we develop a joint approach were state of the art tools from NLP are used to perform both sentiment analysis and identify topics of interest in the students' comments. 

Nonetheless, we must stress that researchers have for long worked on similar problems of assessing customer satisfaction from written opinions including public election forecasting~\cite{smith2010tweets,tumasjan2010predicting,camargo2017}, sales~\cite{liu2007arsa} and trading prediction~\cite{zhang2010trading}, marketing price prediction~\cite{archak2007show}, among others. The common pipeline for performing opinion mining consists of the following general steps~\cite{SUN201710,kumar2016}: \textit{i)} retrieval of opinions from public databases, \textit{ii)} cleaning of the opinions (including discarding some opinions due to quality issues, stemming, tokenisation, among others), \textit{iii)} prediction of a quantity of interest such as polarity, sentiment strength, among others.

In this paper, we combine state-of-the-art methods in an NLP-based pipeline for classifying the sentiment of students' opinions. We then use these results to predict the ratings given to the professors by the students by means of supervised learning algorithms. Furthermore, we perform LDA to discover latent topics that are central in students' opinions. With the power of question answering systems~\cite{ONG2009397}, we envision students' perceptions surveys having only open questions that are fast to answer, reaching high levels of response rates, and also extracting the most relevant information, which comes from the students' opinions. These opinions are then mined with methods like the one we propose to analyse how students truly feel about their professors and classes.

The structure of this paper is as follows. In section methods and materials a brief description of the data and its prepossessing is presented. Next, in the results the analysis of model performance as well as the statistical analysis of the obtained results is scrutinised. Recommendations for future perspectives in the research of the subject as well as an outlined of the conclusions are listed in the final part of the manuscript.

\section{Methods and materials}
\label{sec:methods}

\subsection{Data}

The data used for our study was taken from an anonymised data set from the Konrad Lorenz University in Bogotá, Colombia, which contained around 5,700 professor performance evaluations as perceived by their students. Evaluations from the 2018-2020 period were contained in the data set, accounting for 937 courses (773 undergraduate and 164 graduate courses). The information from the evaluations was separated in two tables. The data included in the first table was:
\begin{itemize}
    \item Subject code: an code that uniquely identifies each subject by year.
    \item Comment: a comment from a student to the professor of the corresponding subject. 
\end{itemize}
The data in the second table was:
\begin{itemize}
    \item Subject code: a code that uniquely identifies each subject by year.
    \item Number of students: the number of students in the corresponding subject.
    \item Professor's pedagogical and disciplinary aspects: a score from 1 to 5. This is an average over all students' evaluations.
    \item Professor's evaluation: a score from 1 to 5 referring to the evaluations carried out by the professor. This is an average over all students' evaluations.
    \item Professor's interpersonal relations: a score from 1 to 5. This is an average over all students' evaluations.
    \item Education level: a binary label taking the values ``undergraduate'' and ``posgraduate'' for the corresponding subject.
\end{itemize}

\subsection{Polarity prediction and topic modelling}
\label{sec:polpred}

\begin{figure}[!ht]
    \centering
    \includegraphics[width=\textwidth]{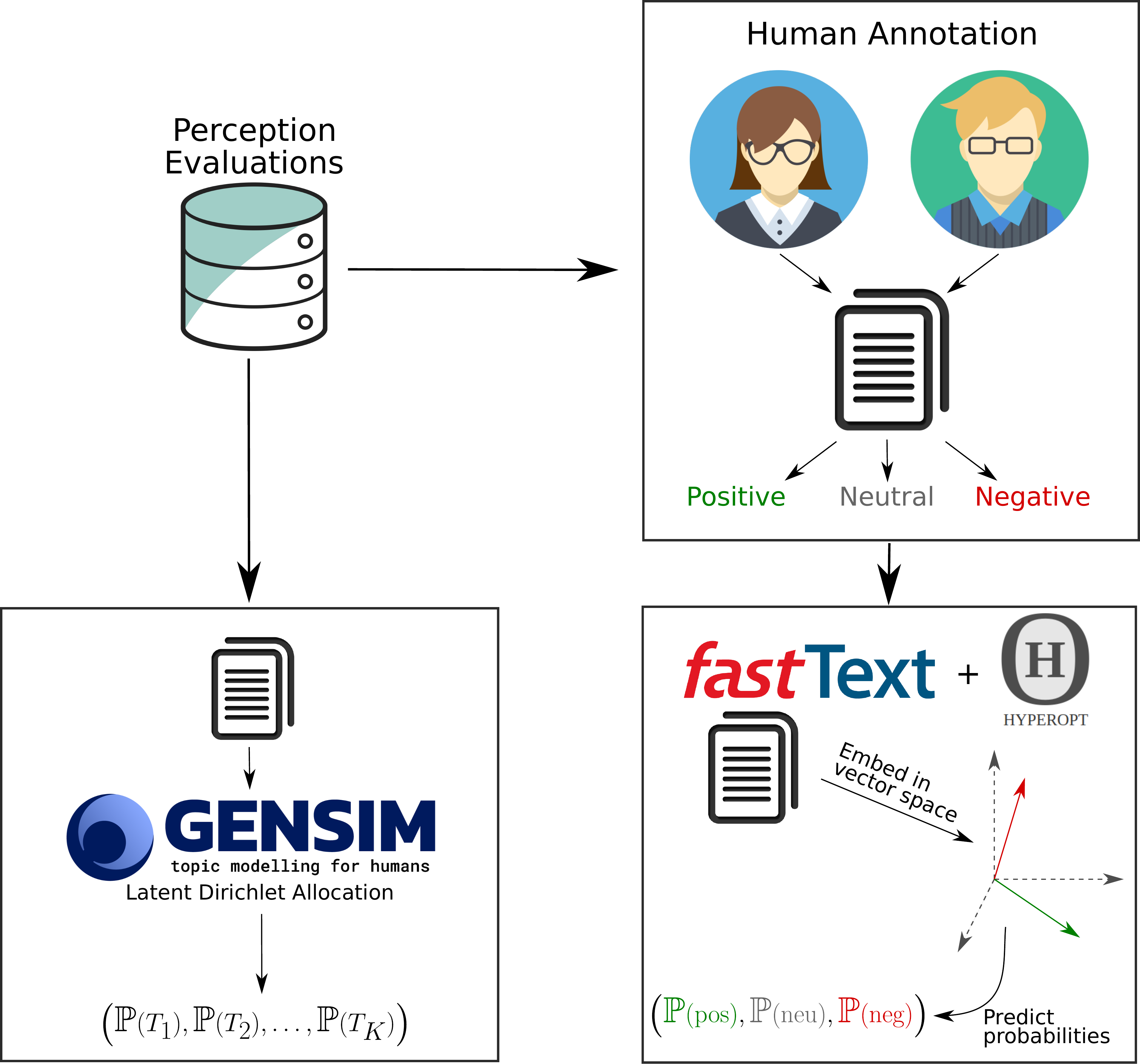}
    \caption{Flow diagram for polarity prediction and topic modelling. First, the raw comments from students are annotated by humans. Then, FastText vectors are obtained for each of them, with a corresponding probability of belonging to one of three polarity classes: positive, negative or neutral. This model's hyperparameters are optimised with Hyperopt. Also, an LDA model from Gensim~\cite{rehurek_lrec} is trained to classify texts into latent topics.}
    \label{fig:flow}
\end{figure}

The methodology for modelling topics in our corpus and for predicting polarity is shown in~\cref{fig:flow}. Not every comment made by students was taken into account. We filtered out those comments with less than 5 words or 10 characters, as they do not contain a lot of information. We ended up with around 4,900 comments that we used for training, validating and testing the methodology hereafter presented. These comments made by undergraduate and graduate students from the years 2018, 2019 and 2020 were annotated by humans with one of three polarity classes: positive, neutral and negative. An example of a positive comment is: ``Thanks for the rigorousness in your subject and for your pedagogy to transmit to us your acquired knowledge''. An example of a negative comment is: ``I suggest the professor to be a bit more organised with respect to time and e-mail reading''.

Each comment from a student passes through a pre-processing stage where stopwords are removed, all characters are lower-cased, punctuation symbols are removed, and words are stemmed. Then, FastText is used to build a polarity classification model. FastText is a natural language processing method to embed text in low-dimensional vector spaces based on the co-occurrence of words within a context~\cite{arm2016bag,bojanowski2016enriching}. The embedding procedure allows FastText to extract latent semantic features encoded into the embedding vector space's dimensions, similar to its predecessor Word2Vec~\cite{le2014distributed}. The polarity classification model is intended to distinguish which regions of the low-dimensional space correspond to one of the three polarities. Because of the low quantity of comments, we selected a total of 20 dimensions to construct the embedding vector space. To tune its hyperparameters we used Hyperopt~\cite{pmlr-v28-bergstra13}, which is a framework that combines randomised search and tree-structured Parzen estimators to optimise an objective function with respect to the FastText hyperparameters. In our case, we measure the quality of classification through the average accuracy defined by
\begin{align}
    S=\frac{1}{3}\sum_{i\in P} a_i, a_i = \frac{\text{\# of comments with polarity $i$ predicted with polarity $i$}}{\text{\# of comments with polarity $i$}}, 
    \label{eq:accuracy}
\end{align}
where $P = \{\text{positive, neutral, negative}\}$. We set the objective function for Hyperopt as
\begin{align}
    {}- S_\text{validation} + \frac{|S_\text{training} - S_\text{validation}|}{1 - S_\text{training} + \epsilon},
\end{align}
where $S_\text{validation/train}$ is the average accuracy for the validation or train sets, and $\epsilon=0.2$ is a positive offset that sets how important it is for Hyperopt to reduce the gap between the accuracy of the training and validation sets. The size of train, test and validation sets were 64\%, 20\% and 16\%, respectively.

Finally, we also trained an LDA model, which is a probabilistic model that assigns each comment to a topic with a probability based on the co-occurrence of words in texts~\cite{blei2003latent}. This allows us to examine if students respond more positively or negatively to different topics. 

\subsection{Prediction of scores from probabilities}

Now, we ask ourselves to which extent can the methodology exposed in~\cref{sec:polpred} be used to predict the numeric score given in professors' performance evaluations. We create a prediction framework where we try to predict numerical scores only from information deduced from the students' comments. This procedure has the potential to give us insights on what is the participation rate of students in open-ended questions along with its relation to the quantitative score given to the course. The prediction is done with XGBoost~\cite{Chen_2016}, which is a widely successful gradient boosting algorithm for regression and classification. We are not interested in precisely predicting the score. Instead, we want to distinguish if students have very high, high or a moderate quality perception of their professor. Therefore, we split the evaluation scores in three groups: very high scores (from 4.5 to 5.0), high scores (from 4.0 to 4.5) and moderate scores (less than 4.0). The classification model takes as an input a FastText vector corresponding to the comments of a class, as well as other features such as the LDA probabilities that those comments belong to one of the $K$ latent topics. Therefore, we use XGBoost to predict, for each class, the average score that students give to their professor based solely on the comments.

For each course we have the average score given by students to that course and the comments registered by students to an open-ended question done at the end of the semester. There may be some students that score the class numerically but do not give any written feedback and vice-versa. This motivates the use of state of the art NLP tools to find out to what extent the average numerical score of a course can be recovered from the comments of the students who took the course. Since there are more courses with high scores than courses with moderate scores we perform a data balancing in order to have the same amount of courses with average grades above and below 4.4, and that balanced data is used to optimise the hyperparameters of the classifier, which in this case is the gradient boosting classifier (XGBoost). After the best hyperparameters were found, the classifier was trained using the complete unbalanced original data.

\section{Results}
\begin{figure}[!ht]
\hfill
\subfigure{\includegraphics[scale=0.39]{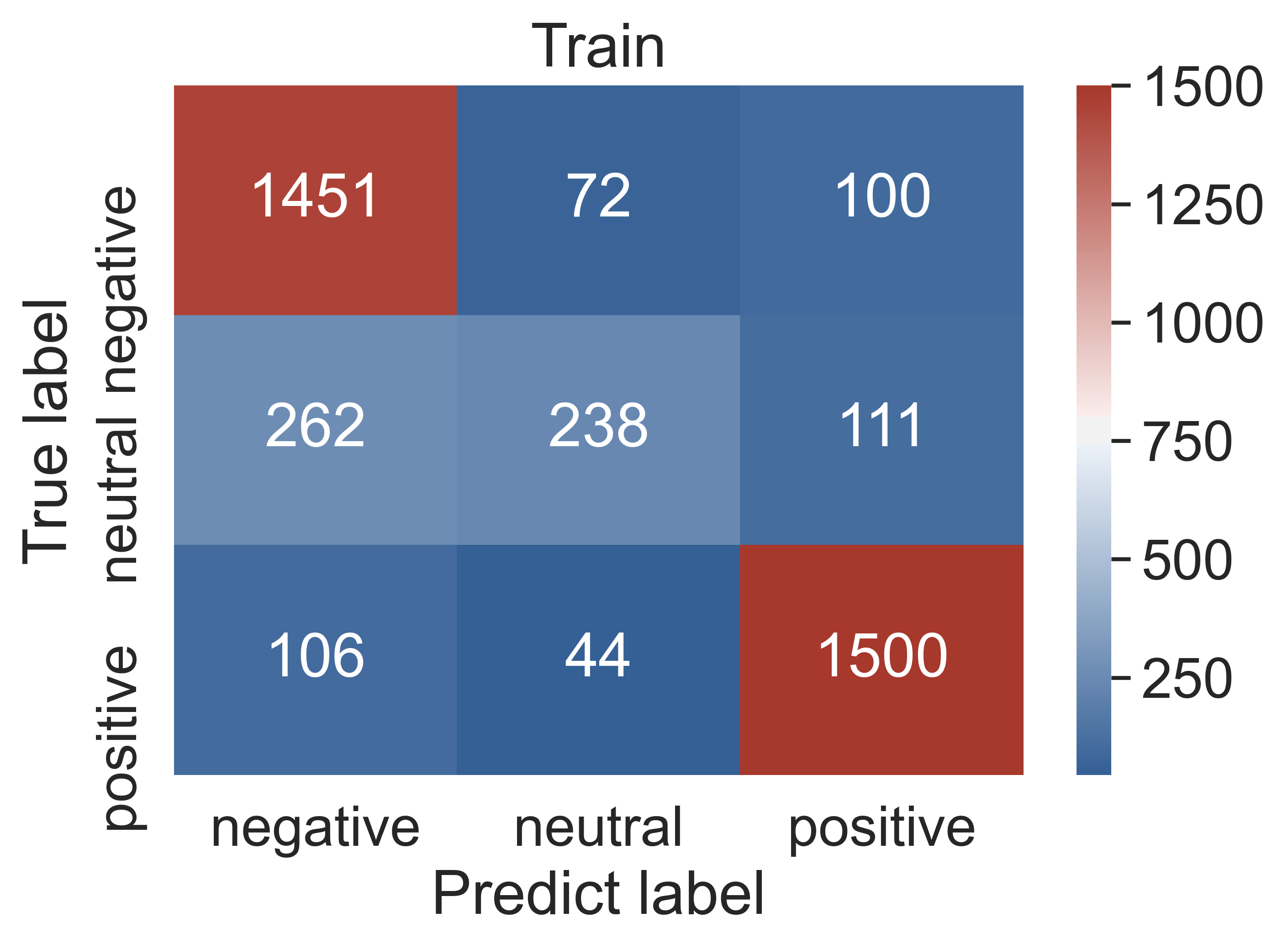}}
\hfill
\subfigure {\includegraphics[scale=0.39]{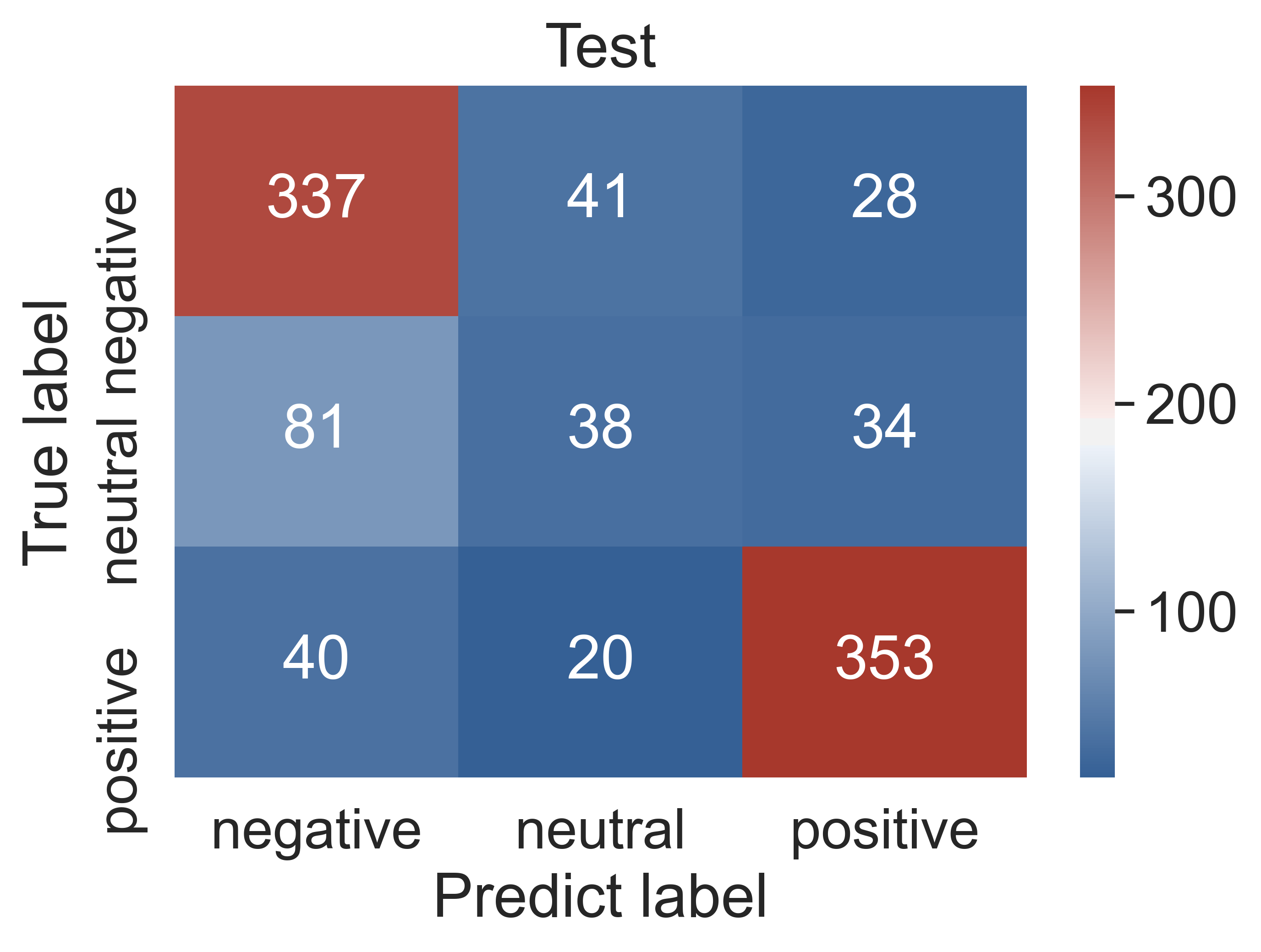}}
\hfill
\caption{Confusion matrices for train and test sets for the FastText model that predicts comment polarity.}
\label{fig:confusionm}
\end{figure}

With the pre-processed data, we trained the FastText model and used Hyperopt to find the best hyperparameters. To measure the precision of this model (see \cref{eq:accuracy}), we use the confusion matrix, which is an error matrix that contains in the diagonal the number of correct predictions for each category, whereas the number of wrong predictions for each category are in the elements outside of the diagonal. Confusion matrices for the train and test sets are shown in \cref{fig:confusionm}. We observed a high number of correct predictions for the categories positive and negative, both in the train and test sets, but a low number of correct predictions in the neutral category, especially in the test set. This can be due to the small number of  neutral comments in the data, or in the difficulty of defining a neutral comment in the annotation process. On the contrary, this effect does not happen for positive and negative comments because the amount of these comments is much higher than the neutral comments. The value of accuracy in the train and test sets is $0.821$  and  $0.749$, respectively, which is stable through the last steps of optimisation, giving evidence that no over-fitting occurs. These results improve over similar pipelines such as the one presented in Ref.~\cite{Camargo_2018}, where high-quality Spanish tweets were annotated by Spanish Society for Natural Language Processing (SEPLN in Spanish). Studies using that dataset (which is similar in data imbalance to ours) also found difficulty in correctly predicting the neutral class~\cite{gonzalez2018elirf,chiruzzo2018retuyt}, probably because of data imbalance, as it was identified in a previous study~\cite{ahpine:hal-01504684}, or because words related to neutral comments might contain sentiment, as indicated in Ref.~\cite{colhon2017objective}. 

\begin{figure}[!ht]
    \centering
    \includegraphics[width=0.7\textwidth]{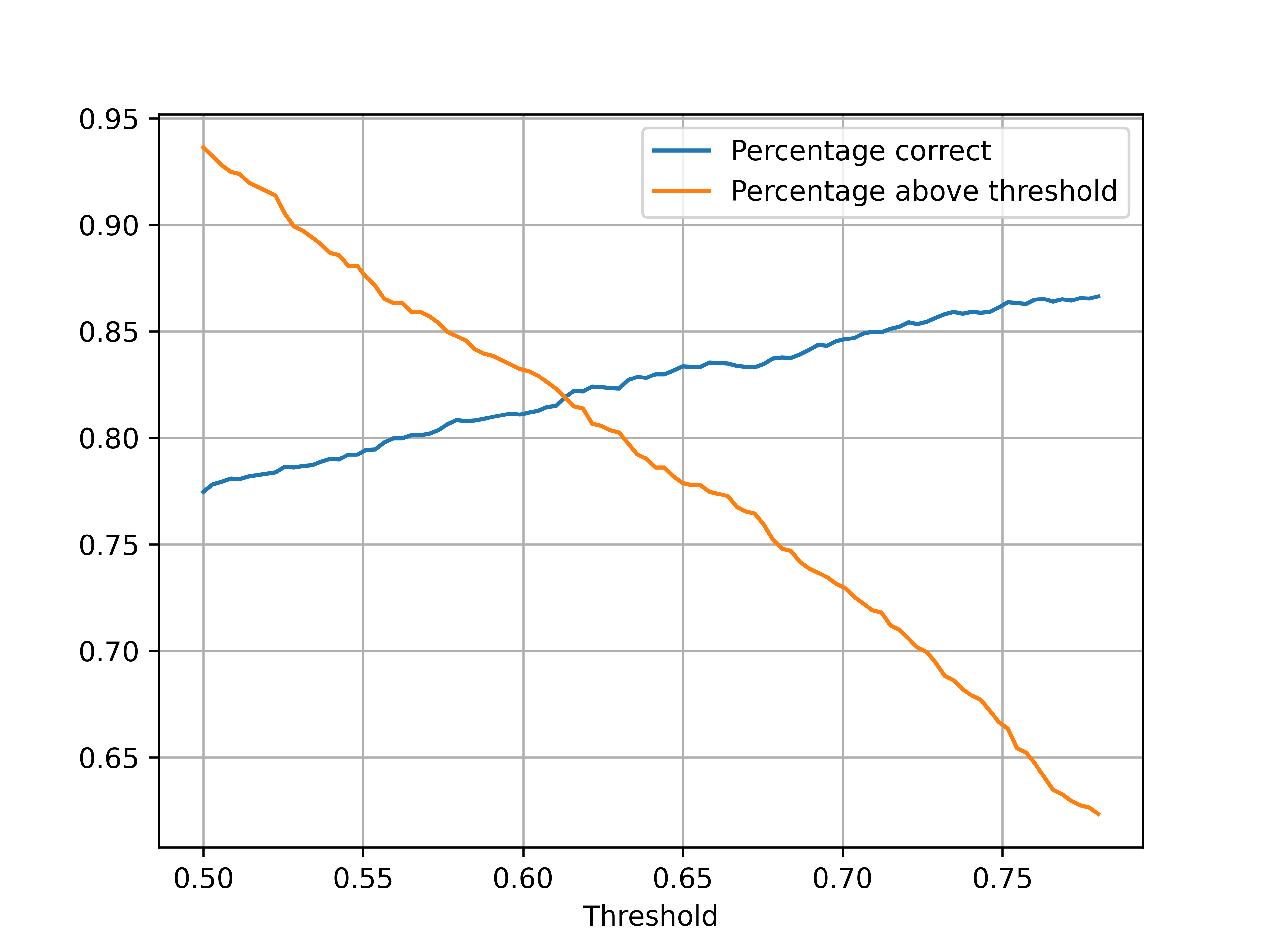}
    \caption{Percentage of correct comments classified with probability greater than a threshold, and percentage of comments with assigned probability greater than a threshold.}
    \label{fig:accuracy}
\end{figure}

Since the FastText model produces the probabilities that a comment is classified as positive, negative or neutral, we can look at the capability of the prediction when the assigned probability is low or high. We analyse this situation in \cref{fig:accuracy}. For a given threshold of probability that the model classifies a comment in a category, the ``percentage above threshold'' is the percentage of comments whose probability is greater than the threshold, while the 
``percentage correct'' is the fraction of comments correctly assigned to a class. We observe, as expected, that the higher the threshold the lower the percentage of comments that have a probability assignation above that threshold. Also, as the assigned probability threshold increases, more comments are correctly classified. These observations are helpful if one wants to make automation decisions about classifying polarity. For instance, an automation rule could be: predict a polarity for a text, if the polarity exceeds a given threshold, take the prediction as the truth, else give the text to a human so that the human classifies it.

For real applications of the model, it is necessary to know the principal topics in the comments with the aim of decision-making within the institutions. We did this with LDA. This model requires a specific number of topics. We found that a number of topics $K=5$ was experimentally good because it allowed us to have good interpretability. Since LDA assigns each comment the probability of belonging to one of the $K$ topics, those comments with high topic probabilities are representative of those latent topics. Reading the representative comments and looking at the most used words in each latent topic, we were able to assign a topic label to each latent topic. The topics are shown in \cref{tab:topics}.
\begin{table}[!ht]
\caption{Topics found with LDA. For each topic we present a relevant negative comment.
\label{tab:topics}}
\centering
\begin{tabular}{c m{0.1cm} m{2.4cm} m{0.1cm} m{2.5cm} m{0.1cm} m{5.7cm}}
\toprule
 \# & & Topic & & Topic Interpretation & & Representative comment\\ 
 \midrule
 0 & & Excellent methodology, practice, dynamic class, school-like, more feedback & & General recommendations about methodology and other aspects of the course & & {``I recommend that the class is taught in the computer room, because the use of EXCEL as a tool was useful to understand the topic when using the formulas and table tools, which could not be done in a classroom without computers.''}\\
1 & & Good professor, dedication, thank you, evaluation criteria, listen to students & & Projects, evaluations and grading schemes & & {``I think that the professor should reestablish her evaluation criteria since she establishes them but she does not accept them when evaluating and this is not fair, since the work is done as she demands, but nothing seems enough for her and she does not value the effort.''} \\
2 & & Best professor, virtual, virtual clasroom, support, feedback & & Aspects of the class methodology  & & {``Pedagogical strategies are good. The time could be distributed to use all the time of the session since the topics are very extensive. In the virtual sessions, it was necessary to use more resources to help students understand, perhaps the board and drawings.''} \\
3 & & Dynamic classes, virtual class, entertaining, class preparation, slow & & Management of time in the projects and class  & & {``The teacher is very well prepared, however, he must control the time available for each activity, since, I give more time to the daily activities but I do not give enough time to the topics and how to develop them in each project.''} \\ 
4 & & Excellent professor, human being, professional, patience, attention & & Professor's attitude and respect towards the class and the students  & & {``The teacher explains the topics of the class clearly, but he lacks courtesy, charisma and decency when addressing his students. From a simple greeting, to respectfully answering a question in the middle of the class.''}\\
\bottomrule
\end{tabular}
\end{table}

\begin{figure}[!ht]
    \centering
    \includegraphics[width=0.6\textwidth]{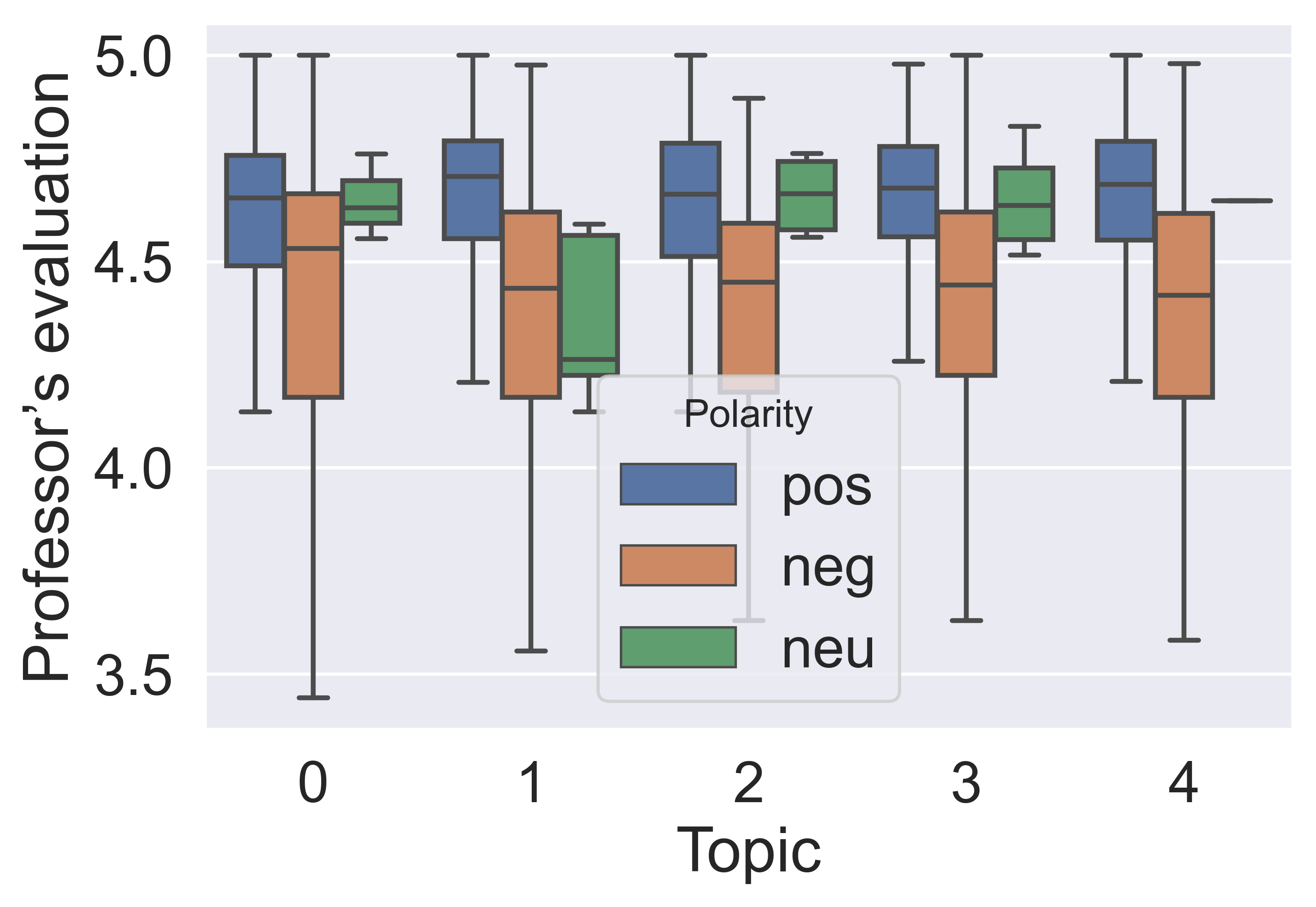}
    \caption{Box plot of the score given by students to professors, grouped by topic and assigned polarity.}
    \label{fig:box_plot_general}
\end{figure}

\Cref{fig:box_plot_general} shows a box plot of the score that students gave to professors, grouped by topic and polarity. We observe that there is a positive correlation between positive/negative polarity (as predicted by the FastText model) and the score given by students. Notice also, that comments with negative polarity have a wider score distribution compared to positive and neutral comments for which the score distributions are narrow and have shorter tails. The mean value in positive and negative comments is almost the same throughout all topics, but for neutral comments, we observed that the mean value is lower in the topic 1. This is because FastText wrongly predicts negative comments as neutral for that topic. This box plot allows us to explore which topics are perceived as better or worst by students when evaluating their professors. In other words, this analysis allows us to know which aspects of professors' teaching generate more discomfort in the students.

One of the most important aspects of surveys is their response rate (RR). We calculated the RR for each course of the university. The value of RR can expose what topic is more relevant for the students positively or negatively. In general the RR in this data is low, but we can do some analysis. For instance, for the topics in \cref{tab:topics}, the highest RR corresponds to topic 0 and the lowest RR to topic 3. This means that the students give recommendations for the methodology of the class, but are more indifferent to the professors' time management. In general, the RR in positive comments is greater than the RR in negative comments, except for topics 1 and 3, where the students tend to have higher RR when expressing their opinions about professors' time management and the way professors evaluate the subjects. On the other hand, the RR is very low in neutral comments for the topics 0 and 1, but these can be biased because of the low quantity of neutral comments in our data set. The results shown above lead us to think that for low RR, we will have low accuracy in polarity prediction. For this reason, it is necessary to encourage the completion of surveys, which leads to the generation of better models and better analysis of the students' comments. 
\begin{figure}[!ht]
    \centering
    \includegraphics[width=0.6\textwidth]{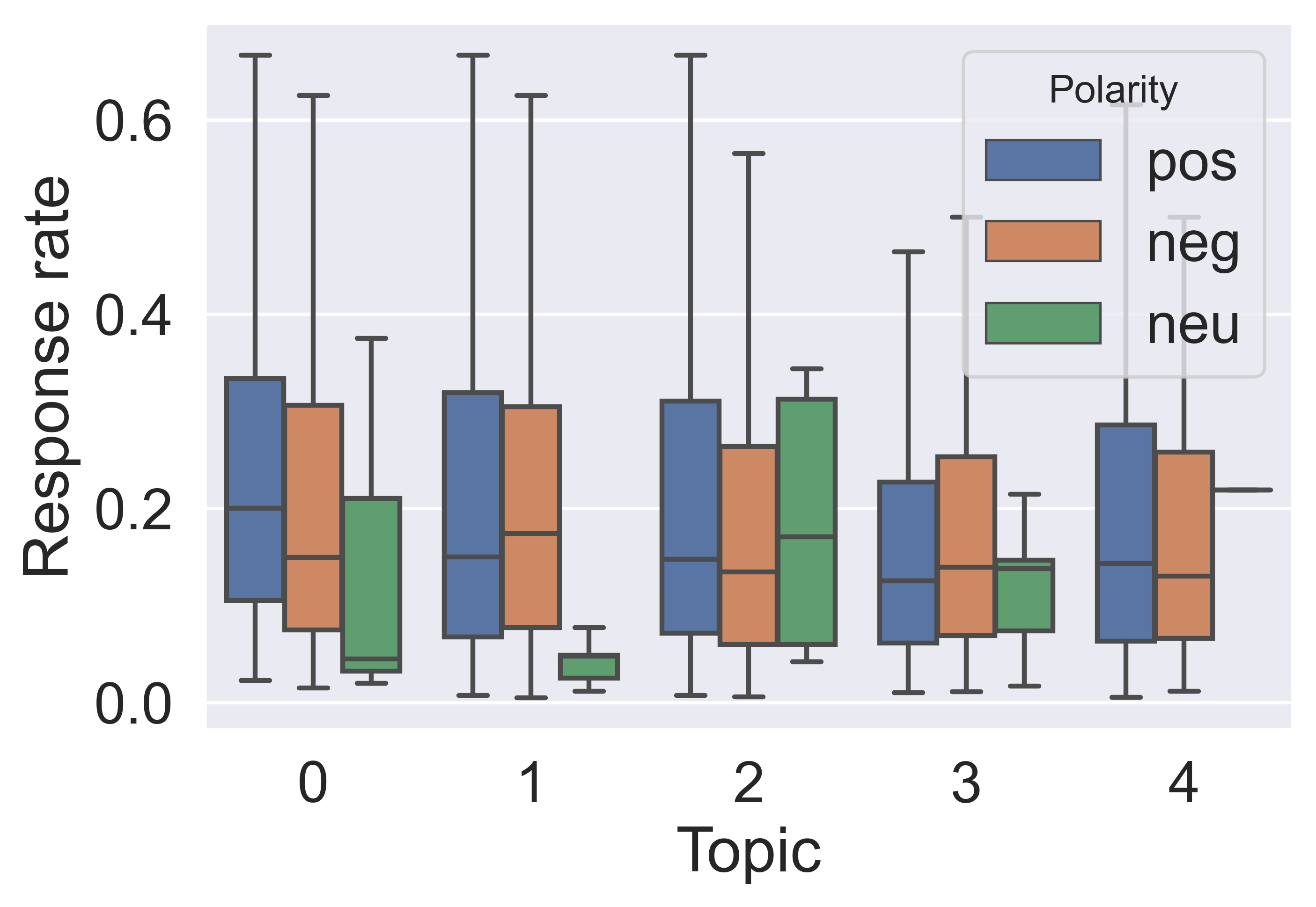}
    \caption{Box plot of the response rate grouped by topic and polarity.}
    \label{fig:box_rr}
\end{figure}

\begin{figure}[!ht]
    \centering
    \includegraphics[scale=0.395]{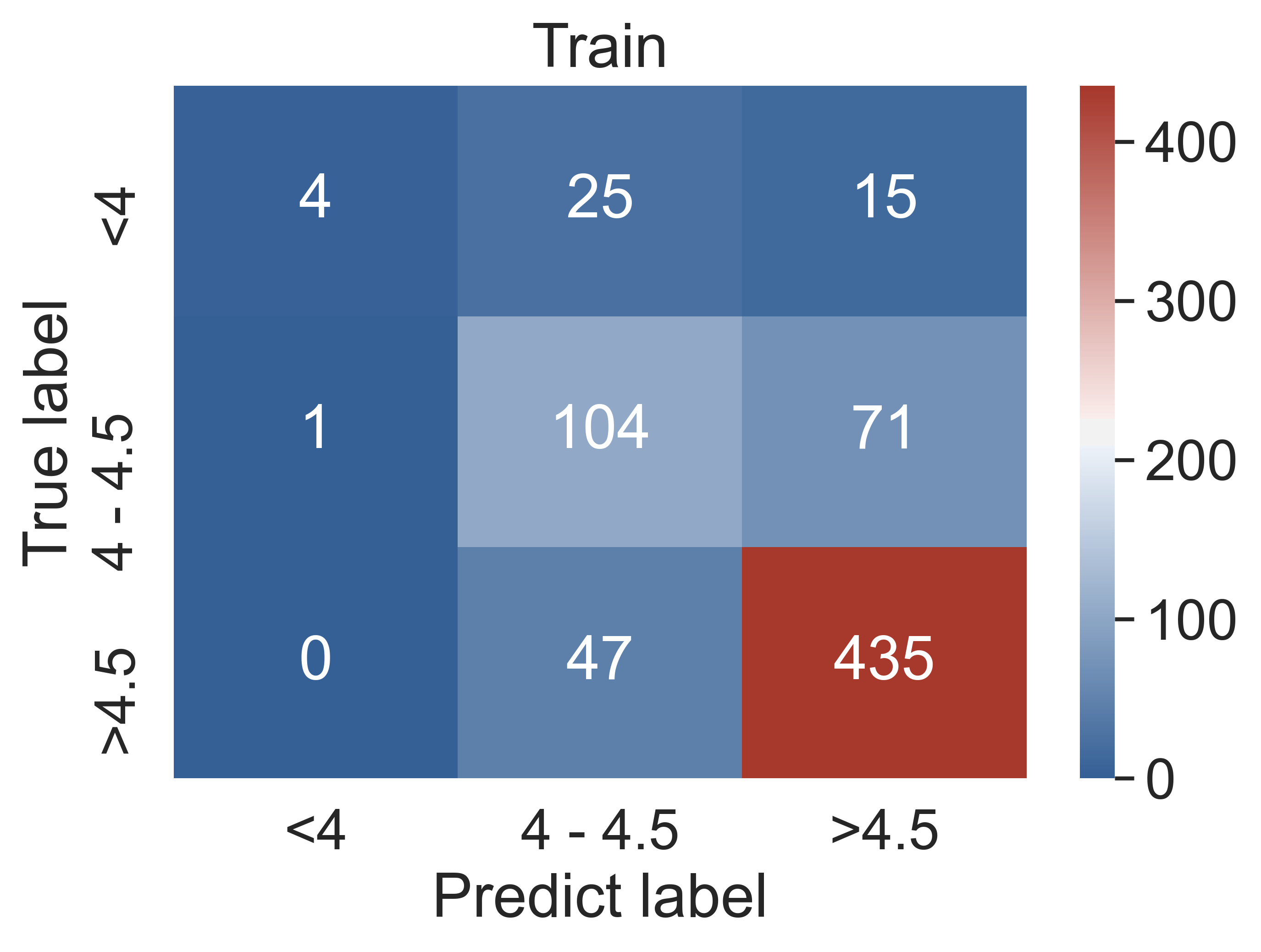}
    \includegraphics[scale=0.395]{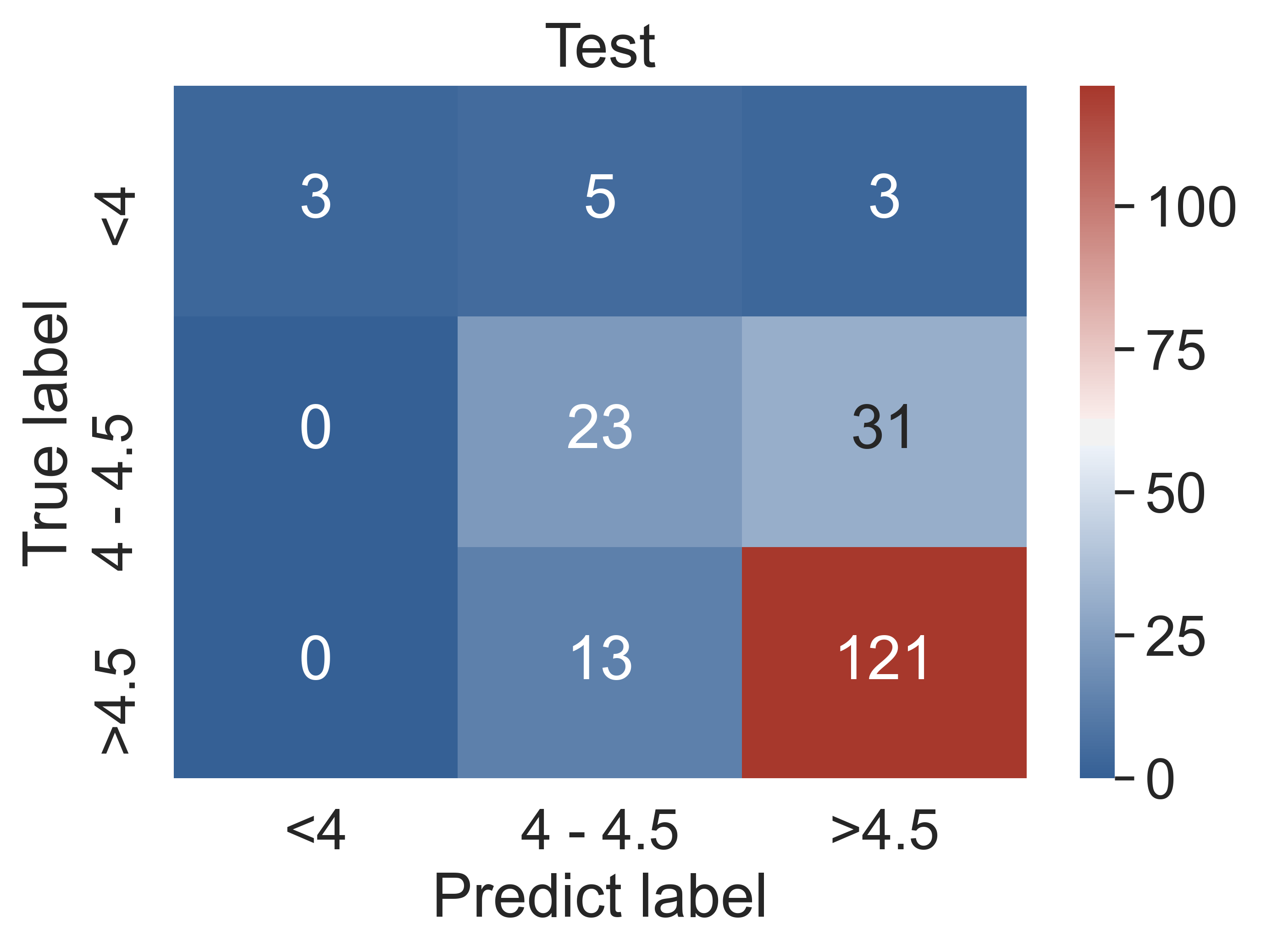}
    \caption{Confusion matrices in train and test sets for the XGBoost score classifier.}
    \label{fig:cmatrix_juanito}
\end{figure}

Finally, we train an XGBoost model to predict the professors' evaluation score intervals (very high, high, and moderate). The confusion matrix for the model is shown in \cref{fig:cmatrix_juanito}. We observed that our model is efficient to predict scores greater than 4.5 (very high scores), but the model fails in moderate scores (less than 4.0). This happens because the database is imbalanced, as it does not have enough comments with moderate scores. The average accuracies of the model, as defined in~\cref{eq:accuracy}, are 0.52 and 0.53 in train and test sets, respectively. Therefore, we conclude that our model does not have a good accuracy to eliminate the closed questions in the professors' evaluation surveys. This is most likely due to the small number of comments in our database. We expect that for a larger comments database, we can build a better model, because it is known that FastText based models (including Word2Vec) have a performance relation with the corpus and vocabulary size~\cite{patel-bhattacharyya-2017-towards,Li2019}.

\section{Limitations and perspectives}

There is room for improvement in some of our methodological steps. We comment below some of the issues that could be addressed in the future in order to improve the accuracies in polarity prediction, as well as in score prediction:

\begin{enumerate}
    \item Improve annotations: It is essential that several experts redundantly label the comments with polarity. Beside the labels used in the present work, we believe two additional labels are of relevance: ``Subjectivity'' and topic.  Subjectivity refers to how much sentiment there is on a comment, regardless of its polarity.
    Topic would help to distinguish between  scenarios, e.g. comments that refer to the professor of the course, to the contents of the course, or to the facilities of the university.
    
    \item Find the optimal number of topics for the LDA topic discovering phase. This can be done through the $C_V$ coherence, which measures how coherent comments belonging to the same topic are~\cite{vargas2020ecuador,roder2015exploring,vargascaldern2019event}.
    
    \item Use pre-trained FastText models reduced to approximately 20 dimensions (the same number of dimensions used in this study), as this may improve the quality of our predictions.
    
    
\end{enumerate}

Furthermore, we shall make explicit some ethical concerns. From an educational perspective, the correct assessment of students' perceptions about their professors and their education is of paramount importance. This assessment allows to take decisions to improve the educational environment, promoting healthier and more productive conditions for students to thrive. However, not only does this assessment affect the lives and projects of the students, but also it affects the professors and universities. Particularly, professors' motivation for delivering high-quality classes can be affected by how their students perceive them. At an institutional level, very important decisions such as removing a professor from a subject, or even firing a professor can be taken using as input the assessment of students' perceptions. Therefore, automated text analysis tools have to be responsively used to assess these perceptions. Our model allows us to take this issue into account, since not only a prediction of the polarity of a comment is given, but also a confidence level over that prediction is also given, as shown in~\cref{fig:accuracy}.

\section{Conclusions}

In this work, we present a new natural language processing methodology to automatically explore students' opinions on their professors, compared to methodologies that use other approaches of sentiment analyses \cite{quteprints115064}. For this, we use state-of-the-art techniques such as FastText to build classifiers that are able to identify polarity in students' opinions. Furthermore, we discover latent topics in the opinions corpus through Latent Dirichlet Allocation. These two approaches are then combined to predict the score that students give to their professors, so that we can identify professors with non-excellent performance only using the information from students' opinions. We argue that such tools can reduce human burden in this analysis, and can also simultaneously take full advantage of the information found in those text opinions. Nonetheless, the experiments so far exposed in our paper indicate that the amount of information or quality of our annotations hampers the possibility of eliminating closed questions from perception surveys to assess the professors' quality.

We envision students' perception questionnaires based on asking the students' for improvement recommendations, and on opinions about negative and positive aspects of the professor and of the class. This questionnaires will probably be more friendly with students, reaching higher response rates, leaving the tedious part of extracting information to the combined work of humans and NLP algorithms, where the heavy-lifting is done by NLP, and only high-level analysis is left for the human. We hope in the future to use larger opinion databases which allow us to find and train a better model, since a larger database enables us to confidently circumvent the imbalance problem through sampling techniques such as the one presented in Ref.~\cite{WANG2013451}.

\section*{Acknowledgements}
V.V., J.F., L.A. and N.P. would like to thank Carlos Viviescas for his useful insights at the beginning of this study.

\bibliographystyle{splncs04.bst}
\bibliography{refs.bib}
\end{document}